\documentclass[10pt,twocolumn,letterpaper]{article}
\usepackage{times,graphicx,amssymb,booktabs}
\usepackage[dvipsnames]{xcolor}
\usepackage[numbers,sort&compress]{natbib}
\usepackage[font=small,labelsep=period]{caption}
\usepackage[letterpaper,textwidth=6.875in,textheight=8.875in,top=1in,columnsep=.3125in]{geometry}
\makeatletter
\renewcommand{\@maketitle}{\begin{center}{\Large\bfseries\@title\par}\vskip 10pt
{\large\begin{tabular}[t]{c}\@author\end{tabular}\par}\end{center}\vskip 8pt}
\renewcommand\section{\@startsection{section}{1}{\z@}{-10pt plus -2pt minus -2pt}{7pt}{\large\bfseries}}
\renewcommand\subsection{\@startsection{subsection}{2}{\z@}{-8pt plus -2pt minus -2pt}{5pt}{\fontsize{11}{13}\selectfont\bfseries}}
\makeatother
\renewenvironment{abstract}{\centerline{\large\bfseries Abstract}\vspace{12pt}\par\noindent\itshape}{\par\vspace{12pt}}
\ifdefined\XeTeXversion
  \usepackage{fontspec}
\fi
\usepackage{mathtools}
\newcommand{\method}{RelPRM}
\newcommand{\trainset}{RelProcessTrain}
\newcommand{\benchset}{RelProcessBench}
\newcommand{\sg}{\operatorname{sg}}
\newcommand{\KL}{D_{\mathrm{KL}}}
\newcommand{\SL}{\operatorname{SmoothL1}}
\newcommand{\ind}{\mathbf{1}}

\usepackage{pdflscape}

\usepackage{colortbl}
\newcommand{\resultgroup}[2]{\rowcolor[gray]{.92}\multicolumn{#1}{c}{\small\bfseries #2}\\}

\usepackage{multicol}

\definecolor{linkblue}{rgb}{0.21,0.49,0.74}
\usepackage[pagebackref,breaklinks,colorlinks,allcolors=linkblue]{hyperref}
\title{RelPRM: Grounding Visual Process Reward Models\\in Relational Evidence}
\author{Zhaoxing Zhang\textsuperscript{1}\qquad Ziqiu Gui\textsuperscript{2}\qquad Cheng Chen\textsuperscript{2}\\[3pt]
\normalsize\textsuperscript{1}Huazhong University of Science and Technology\\
\normalsize\textsuperscript{2}Xiaohongshu}
\date{}
\hypersetup{pdfauthor={Zhaoxing Zhang, Ziqiu Gui, Cheng Chen},pdftitle={RelPRM: Grounding Visual Process Reward Models in Relational Evidence}}
\begin{document}
\maketitle
\begin{abstract}
Visual process reward models (PRMs) can detect calculation errors yet assign high scores to steps that misread visual relations. Supervision based on subsequent answer success does not directly establish whether a step's relational claim holds. To strengthen visual-relation verification, we first build an automated pipeline that programmatically verifies each claim and identifies the required objects, relations, and image regions. It produces \trainset{} and \benchset{} (RPB), which tests whether scores respond appropriately to relational evidence through controlled image edits and step rephrasings. Building on this pipeline, we further introduce \method{}, a visual-relation-oriented method for process reward modeling with two training mechanisms. Relation-evidence self-distillation constructs views preserving necessary objects, relations, and dependencies, reducing the burden of finding decisive evidence in complex scenes and transferring reliable evidence-assisted judgments to full images. Relation-evidence-driven consistency training encourages appropriate score changes when relation edits reverse a step's truth, stability under irrelevant edits, and similar score changes across paraphrases. At inference, \method{} directly scores the current step using only the original image and reasoning context. RelPRM achieves 66.2\% on VisualProcessBench (VPB) and 68.4\% on RPB, gains of 4.2 and 9.9 percentage points over VisualPRM-8B. These results show that training for visual relations improves both relational verification and step assessment across broader multimodal reasoning tasks.
\end{abstract}

\section{Introduction}
\label{sec:intro}
\begin{figure}[t]
\centering
\includegraphics[width=\columnwidth]{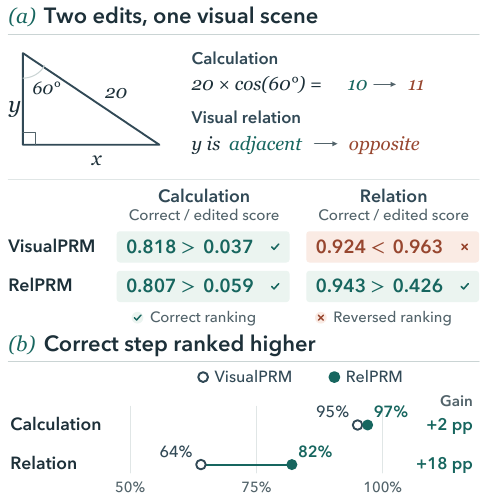}
\caption{\textbf{RelPRM corrects a visual-relation ranking.} (a) Both models favor the correct calculation. Only RelPRM favors the correct description of side $y$. (b) On 100 paired questions, RelPRM ranks 18 more relation pairs and two more calculation pairs correctly than VisualPRM. Ties count as nonwins.}
\label{fig:diagnostic}
\end{figure}

A visual process reward model (PRM) scores the correctness of the current reasoning step given an image, a question, and the preceding steps. These scores help locate reasoning errors and rank candidate solutions~\cite{verify,visualprm,mmprm,ong}. In visual reasoning, a step's correctness depends on both the validity of its calculation and the accuracy of the image relations on which it relies.

However, existing models differ markedly in their ability to judge calculations and visual relations. In the triangle in Fig.~\ref{fig:diagnostic}, changing $20\cos(60^\circ)=10$ to $20\cos(60^\circ)=11$ lowers VisualPRM's score from 0.818 to 0.037. Calling the leg adjacent to the marked angle opposite instead raises its score from 0.924 to 0.963. To quantify this difference, we construct matched pairs on 100 questions from VisualProcessBench (VPB). For each question, we select a calculation step and a visual-relation step, then construct correct and erroneous versions of each, yielding 400 fixed inputs. VisualPRM ranks the correct version higher in 95\% of calculation pairs, but in only 64\% of visual-relation pairs.

One source of this gap is that existing training-data generation cannot guarantee the quality of visual-relation supervision. VisualPRM400K, a PRM training dataset, is generated using Monte Carlo rollouts~\cite{visualprm}. Prior work shows that Monte Carlo rollout-based labels estimate downstream answer success, which may diverge from current-step correctness~\cite{lessons,beyondfirst,noiseaware}. For example, an arrow may point from A to B while a step states that B points to A. Subsequent reasoning can bypass this error and still reach the correct answer. Visual relations require direct verification of the step by locating the visual information and determining its direction. We therefore investigate supervision with verifiable relational truth, corresponding visual evidence, and controlled relation changes. Our \textbf{RelProcess} pipeline programmatically identifies the objects, relations, and dependencies required by each step, then rechecks the same statement on edited scenes. It produces \trainset{} and \benchset{} (RPB), with evaluation splits for new scenes, unseen wording, and unseen task compositions.

Training on \trainset{} improves visual-relation verification, yet labels alone remain insufficient with more objects, similar distractors, and deeper necessary dependencies. Relation-data training still drops substantially on RPB's Hard subset (Appendix~\ref{sec:difficulty-analysis}). We propose \textbf{\method{}}, a visual-relation-oriented training framework with two mechanisms for using necessary evidence. \emph{Relation-evidence self-distillation} reduces the burden of finding and distinguishing necessary evidence while preserving the task~\cite{visionopd,opdv}. An evidence view suppresses distractors while retaining necessary dependencies, including all nodes and edges needed for a path judgment. The same model scores the same step on both views. Its evidence-view prediction guides the full-image judgment only when it predicts the verified label and assigns that label a higher probability than the full-image prediction does. This transfers reliable evidence-assisted judgments to the original input. \emph{Relation-evidence-driven consistency training} holds a step statement fixed and verifies it on the original and edited scenes. If moving an object reverses a position claim's truth, the version in which the claim holds should receive a higher score. If the edit preserves an earlier edge observation and its supporting evidence, that step's score should stay stable. Paraphrases of the same claim should undergo similar score changes under the same edit. At inference, \method{} scores each step in one pass using only the original image and reasoning context.

The trained RelPRM corrects the visual-relation ranking in Fig.~\ref{fig:diagnostic}, scoring the correct description 0.943 and the erroneous description 0.426, while also ranking the correct calculation above its erroneous version. On the 100-question diagnostic, visual-relation ranking accuracy rises from 64\% to 82\%, and calculation ranking accuracy from 95\% to 97\%. The gap narrows from 31 to 15 percentage points, with details in Appendix~\ref{sec:diagnostic-appendix}. This improvement also extends to broader benchmarks. RelPRM-8B reaches 68.4\% on RPB, exceeding VisualPRM-8B by 9.9 percentage points (Table~\ref{tab:relations}). On VPB, a benchmark of general multimodal reasoning, it also gains 4.2 percentage points (Table~\ref{tab:vpb}). These results show that training for visual-relation verification can also improve step assessment across a broader range of multimodal reasoning tasks.

\section{Related Work}
\label{sec:related}
\begingroup
\renewcommand{\paragraph}[1]{\par\addvspace{8pt}\noindent\textbf{#1}\enspace\ignorespaces}
\paragraph{Multimodal process reward models.}
Process supervision judges intermediate steps to locate errors and select solutions. Human and automatic labels underpin mathematical verification~\cite{verify,mathshepherd}. Multimodal training requires scalable labels. URSA, VisualPRM, and MM-PRM address this through automatic supervision, training data and evaluation resources, and tree-search-based annotation~\cite{ursa,visualprm,mmprm}. Research also examines visual evidence. Perception-focused supervision targets image-reading errors~\cite{ong}, generative analyses and chain-of-thought strengthen verification~\cite{gmprm,vrprm}, and tools or structured constraints support judgments~\cite{timprm,evpv}. ProcessBench, PRMBench, and ViLBench assess error localization, sensitivity, and process judgments~\cite{processbench,prmbench,vilbench}. RelPRM advances visual verification by jointly training on relational truth, required evidence, and evidence changes for direct full-image scoring.

\paragraph{Verifiable visual-relation supervision.}
Structured scenes and executable programs make relations directly checkable. CLEVR and GQA use scene representations and programs for compositional visual questions, while SpatialSense targets relations difficult to infer from object identities~\cite{clevr,gqa,spatialsense}. Applying this approach to process supervision requires connecting facts to individual steps. PaLMR builds alignment signals from verifiable visual facts and structured pseudo-ground-truths, guiding a generation model through process rewards~\cite{palmr}. RelProcess connects each claim, its necessary evidence, and image edits for a step scorer. With prefix and statement fixed, program re-execution determines whether an edit changes step truth or required evidence. Moving a reference object can reverse a position claim yet preserve an earlier edge observation. This step-level correspondence supplies correctness labels and edit-response supervision.

\paragraph{Evidence-assisted training and consistency.}
Distillation enables learning from additional training-time information~\cite{kd,privdist}. Vision-OPD uses cropped views, while OPD-V and visually attributable self-distillation select guidance through visual contrasts~\cite{visionopd,opdv,vag}. These motivate transferring evidence-assisted judgments to full images. RelPRM constructs views from program-recorded dependencies and selects reliable teacher predictions using verified labels. Evidence use also requires appropriate edit responses. Paired tests expose sensitivity to images and wording, while Crome trains rewards with targeted and neutral augmentations~\cite{vqav2,winoground,aro,sugarcrepe,eqben,crome}. RelPRM uses step dependencies to learn truth-change ranking, irrelevant-edit stability, and consistent responses across paraphrases.
\par\endgroup

\section{RelProcess: Verifiable Relational Process Data}
\label{sec:data}
\begin{figure*}[t]
\centering
\includegraphics[width=.90\textwidth]{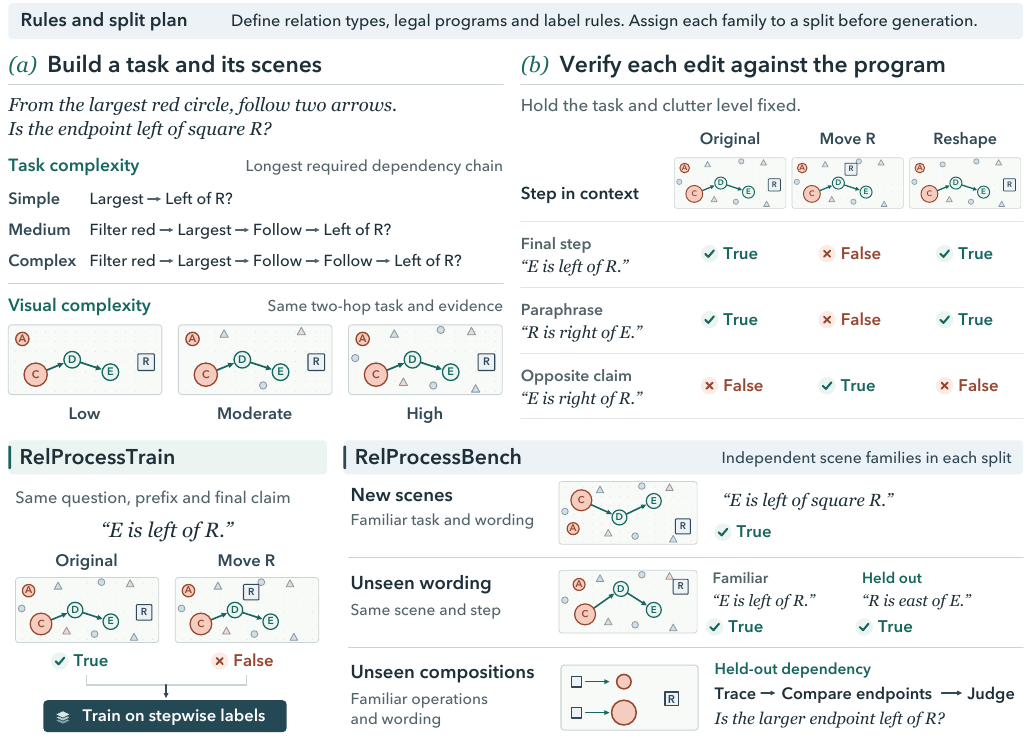}
\caption{\textbf{RelProcess data construction.} Families are assigned to splits before generation. (a) Task depth follows necessary program dependencies, while clutter varies around the same required evidence. (b) Each fixed claim is checked against original and edited scenes: moving R flips the spatial claim, whereas reshaping a distractor preserves it. Correct and error-containing trajectories receive step labels, and equivalent expressions retain correspondence. Evaluation separates new scenes, held-out wording, and held-out typed programs.}
\label{fig:data}
\end{figure*}

\subsection{Executable tasks and step labels}
\begin{table}[t]
\centering\small
\setlength{\tabcolsep}{3pt}
\begin{tabular}{p{.24\columnwidth}p{.69\columnwidth}}
\toprule
Relation group & Subtypes\\
\midrule
Spatial & S1 Axis position, S2 Between, S3 Containment, S4 Overlap\\
Comparison & C1 Segment length, C2 Filled area, C3 Conditional count, C4 Chart values\\
Ordering & O1 Global rank, O2 Conditional rank, O3 Rank adjacency, O4 Two-key ordering\\
Graph & G1 Direct edge, G2 Reachability, G3 Exact hop count, G4 Shortest path\\
\bottomrule
\end{tabular}
\caption{\textbf{RelProcess taxonomy: four groups, four subtypes per group, 16 in total.} Each subtype ID combines the group initial with a number from 1 to 4.}
\label{tab:taxonomy}
\end{table}

A data item contains an image $I$, question $q$, trajectory $(s_1,\ldots,s_K)$, and labels $(y_1,\ldots,y_K)$. A \emph{scene family} groups an independent base scene and task with its clutter levels, edits, error trajectories, paraphrases, and evidence views. Families and all derivatives are assigned to one training or evaluation split before generation; reused base sources remain together.

RelProcess covers four relation groups: Spatial, Comparison, Ordering, and Graph. Each group has four subtypes, giving 16 in total (Tab.~\ref{tab:taxonomy}). It first samples a legal task program and its dependencies, then generates a structured scene with object identities, attributes, positions, connections, or numeric values (Fig.~\ref{fig:data}(a)). Executing the program yields intermediate results and a final answer. A renderer produces the corresponding image; the construction metadata remains outside the scoring prompt.

Rendering and verification share scenes, aligning textual references with image objects and attributes. Labels follow each step's relations and operations rather than final-answer success alone, supporting evidence and dependency recording.

For each image, we construct a correct trajectory and a trajectory containing an injected error, then render two equivalent textual versions. Errors occur at intermediate as well as terminal positions. Labels are recomputed for each step instead of marking every subsequent step as incorrect. In particular, the verifier resolves references using the values asserted in the preceding trajectory, checks a current operation locally, and separately checks terminal agreement with the independently executed task answer. A locally valid intermediate operation after an erroneous premise may therefore receive a positive label; dependency and error-origin metadata preserve that distinction. Appendix~\ref{sec:relprocess-examples} shows one example from the new-scenes split for each of the 16 subtypes, with a labeled trace and a paired incorrect step (Figs.~\ref{fig:plate-spatial}--\ref{fig:plate-graph}).

\newpage
\subsection{Separating task depth from visual clutter}
\label{sec:complexity}
Task complexity is defined by the longest necessary dependency chain of a program, after fixing the granularity of primitive operations. The three bands contain 2, 3--4, and 5--6 necessary operations. We also retain total operation count, branches, comparison scope, and textual length. Parallel reads can yield a written trajectory longer than its longest dependency chain. The current release contains trajectories of 2--9 steps.

Visual complexity controls content outside the required evidence. The current generator adds 1, 3, or 7 distractor objects, varying their similarity to the relevant objects. For each task, its three visual versions preserve the necessary evidence and verified labels. Added candidates or edges are not assumed irrelevant: the program must recheck whether they change a selection, comparison, or traversal. Varying clutter at each depth yields a $3\times3$ grid. Fig.~\ref{fig:data}(a) shows a two-hop task with the same evidence at three clutter levels. Subtypes are assigned only to naturally supported configurations.

\subsection{Edits, paraphrases, and evidence views}
At each clutter level, the generator creates an original image, a target edit, and an irrelevant-content edit (Fig.~\ref{fig:data}(b)). Every edit is re-executed against the step dependencies. An edit may change one step and preserve another; its name alone never determines a label. The target-edited image has its own recomputed trajectories, and a separate cross-image label matrix records the truth of the \emph{same fixed statements} on the different images. This matrix is necessary for consistency learning: two independently regenerated correct answers do not form a true--false pair.

Two expressions share a semantic trajectory, step boundaries, and labels. Opposite claims are distinguished from equivalent ones. For example, ``E is left of R'' and ``R is right of E'' are paraphrases, whereas ``E is right of R'' has the opposite truth condition. Controlled grammar provides reproducible wording splits, although it does not represent all forms of natural-language variation.

The program also records the evidence needed for a current step and its dependencies. This includes intermediate path nodes and edges, comparison candidates, and necessary reference objects. A training-time evidence view can attenuate distractors or isolate these elements while preserving their relative geometry. A view that removes competing candidates from an extremum judgment is invalid even if it looks simpler. Evidence views contain neither correctness labels nor answer overlays.

\subsection{Release statistics and evaluation splits}
\begin{table}[t]
\centering\small
\setlength{\tabcolsep}{2pt}
\begin{tabular}{lrrrr}
\toprule
Split & Families & Images & Trajectories & Labels\\
\midrule
Train & 12,000 & 108,000 & 432,000 & 1,666,296\\
New scenes & 900 & 8,100 & 32,400 & 124,956\\
Unseen wording & 900 & 8,100 & 32,400 & 124,956\\
Unseen comp. & 900 & 8,100 & 32,400 & 134,604\\
\midrule
Bench total & 2,700 & 24,300 & 97,200 & 384,516\\
\bottomrule
\end{tabular}
\caption{Verified construction statistics for the current RelProcess release. Trajectories count both expressions; step counts include positive and negative labels.}
\label{tab:data}
\end{table}

The current verified release contains 12,000 training families and 2,700 evaluation families (Tab.~\ref{tab:data}). Each family contributes three clutter levels, three image variants per level, and two semantic trajectories with two expressions each: nine images and 36 expressed trajectories. Step counts are summed from the actual trajectories, rather than estimated using a fixed length.

RPB has three 900-family conditions (bottom of Fig.~\ref{fig:data}). \textbf{New scenes} use independent instances with familiar wording and task structure. \textbf{Unseen wording} pairs a familiar expression with a held-out expression on the same new scene; the held-out expression must be evaluated separately. \textbf{Unseen compositions} use familiar primitives and wording with held-out full typed programs. Of these 900 families, 824 have unseen operation topologies and 76 have unseen typed attribute configurations. 

Internal checks verify family isolation, image uniqueness, program execution, label alignment across paraphrases and clutter levels, and held-out composition separation.

\section{RelPRM Training Framework}
\label{sec:method}
\begin{figure*}[t]
\centering
\includegraphics[width=.90\textwidth]{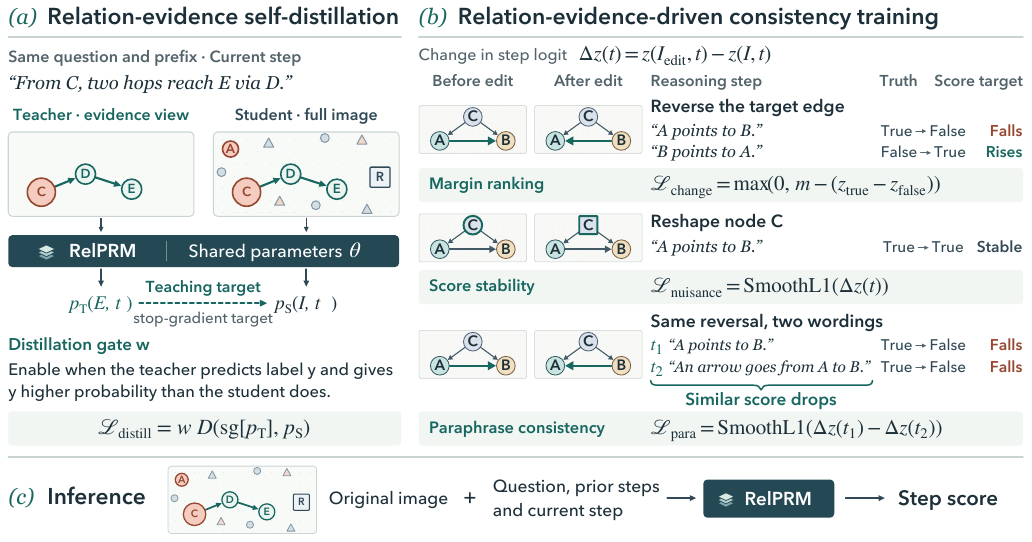}
\caption{\textbf{RelPRM training and inference.} (a) Two shared-parameter passes receive the same question, prefix, and current step. The teacher view preserves the complete required path. Full-image GT supervision is always retained; detached distillation is added only when the teacher is correct and assigns more probability to the verified class $y$. The teacher $p_T(E,t)$ sees evidence view $E$; the student $p_S(I,t)$ sees full image $I$, with shared text context $t$. (b) Program-verified edits define truth-change ranking, nuisance stability, and agreement between edit-induced logit changes across paraphrases. (c) Inference uses the full image and reasoning context in one scorer pass.}
\label{fig:training}
\end{figure*}

\subsection{Step scoring and ground-truth supervision}
Let $t_i=(q,s_{<i},s_i)$ denote the question, the preceding steps, and the current step. The scorer receives $(I,t_i)$ and produces a native correctness logit $z_\theta(I,t_i)$. When the interface uses positive and negative verdict tokens, $z$ is their logit difference. The normalized probability is $p_\theta(I,t)=\sigma(z_\theta(I,t))$. No later steps, program answers, or ground-truth labels are included in $t_i$.

Every verified input used for supervised training retains its own label $y\in\{0,1\}$. The basic loss is
\begin{equation}
\ell_{\mathrm{GT}}(I,t,y)=-y\log p_\theta-(1-y)\log(1-p_\theta).
\label{eq:gt}
\end{equation}
We average within a trajectory and then across sampled trajectories or families so that longer trajectories do not automatically dominate. A ground-truth-only control uses the same labels and paired inputs as the method.

\subsection{Relation-evidence self-distillation}
\label{sec:resd}
A binary step label tells the verifier whether a claim holds. It gives no direct indication of where to look. To supervise the full-image judgment, we ask the same model to check the claim on a second view that retains its supporting evidence. We use this prediction as a teaching signal only if the evidence view improves the model's judgment of the verified class.

\paragraph{A verified view of the same judgment.}
The student sees the full image $I$, while the teacher pass sees a verified evidence view $E_i$ (Fig.~\ref{fig:training}(a)). Both receive exactly the same $t_i$ and label semantics. Supplementary Fig.~\ref{fig:evidence-example} shows the full image and both evidence views for the same step. The view reduces the search and discrimination burden without removing a required reasoning operation. It can be constructed at any clutter level: a nominally simple scene may still benefit from evidence assistance. Conversely, lower clutter does not guarantee a better teacher prediction.

Both passes use the same current parameters $\theta$. Define the binary distributions
{\small
\begin{equation}
P_S=(1-p_\theta(I,t),p_\theta(I,t)),\quad
P_T=(1-p_\theta(E,t),p_\theta(E,t)).
\end{equation}}

The teacher prediction supplies an auxiliary target only when it predicts the verified class and supports that class more strongly than the student:
\begin{equation}
w=\ind\!\left[\arg\max_c P_T(c)=y\right]
\ind\!\left[P_T(y)>P_S(y)\right].
\label{eq:gate}
\end{equation}
We stop gradients through the gate and the teacher prediction. For an incorrect step ($y=0$), the gate compares the probabilities assigned to the negative class. Both passes can already predict the correct label and still satisfy the gate. Every full-image input retains its supervised loss in Eq.~\eqref{eq:gt}, including when the gate is closed.

For a batch $\mathcal B$ of verified candidate teaching pairs, we use
\begin{equation}
\mathcal L_{\mathrm{distill}}=
\frac{1}{|\mathcal B|}\sum_{j\in\mathcal B}
w_j\,\KL\!\left(\sg[P_{T,j}]\,\Vert\,P_{S,j}\right).
\label{eq:distill}
\end{equation}
We normalize by the number of candidate pairs. The effective weight of this loss therefore depends on how often the gate opens. Only the full-image pass receives gradients from distillation. The two passes share weights, so each update also changes the model used to produce the next teacher prediction. Table~\ref{tab:components} compares the gate, teacher update scheme, and evidence-view supervision.

\subsection{Relation-evidence-driven consistency training}
\label{sec:consistency}
Consider moving a reference square R. The edit may reverse the claim that E lies to the left of R and leave the observation that C points to D intact. One image edit thus calls for different score changes at different steps. If two steps express the same claim in different words, their scores should respond similarly to the move.

For an image pair $(I,I')$, we hold the question, prefix, and current claim $t$ fixed. The program checks the truth and evidence dependencies of that claim on both images. These checks determine which pairs contribute to the three losses in Fig.~\ref{fig:training}(b).

\paragraph{Truth-changing edits.}
If the same step has opposite labels in the two images, let $z_+$ and $z_-$ be the logits on its true and false versions. We impose a finite margin:
\begin{equation}
\ell_{\mathrm{change}}=\max\!\left(0,m-(z_+-z_-)\right),\quad m>0.
\label{eq:change}
\end{equation}
The true version can be either the original or the edited image. The loss stops increasing separation once the margin is reached. Each version also receives its own GT label, since correct ranking alone does not establish correct absolute judgments.

\paragraph{Irrelevant edits.}
An edit qualifies for stability only when it preserves the current step's meaning, evidence, and necessary dependencies. Equal labels alone are insufficient. Define the response using the sign convention in Fig.~\ref{fig:training}(b):
\begin{equation}
\Delta z(t)=z_\theta(I',t)-z_\theta(I,t).
\label{eq:delta}
\end{equation}
The stability loss is
\begin{equation}
\ell_{\mathrm{nuisance}}=\SL\!\left(\Delta z(t)\right).
\label{eq:nuisance}
\end{equation}
It permits small fluctuations and applies to eligible negative as well as positive steps. Moving R in a position question, for instance, can be irrelevant to an earlier edge observation but relevant to the final spatial claim.

\paragraph{Equivalent wording, consistent responses.}
For two verified paraphrases $t_1,t_2$ under the same truth-changing edit, we align their logit changes:
\begin{equation}
\ell_{\mathrm{para}}=\SL\!\left(\Delta z(t_1)-\Delta z(t_2)\right).
\label{eq:para}
\end{equation}
This allows different base confidences while discouraging wording-dependent reactions to the same evidence change. A pair of constant scores would satisfy this term, so both expressions retain GT and, when eligible, the truth-change margin. Opposite claims are supervised with their own labels instead of being treated as paraphrases. Unlike the detached teacher in RESD, both endpoints of each consistency constraint receive gradients.

\subsection{Combined objective and inference}
Each consistency term is averaged over its own eligible pairs, with zero contribution when none are available. This prevents numerous unaffected steps from overwhelming scarce truth-changing ones. The combined objective is
\begin{equation}
\begin{split}
\mathcal L={}&\mathcal L_{\mathrm{GT}}
+\lambda_d\mathcal L_{\mathrm{distill}}
+\lambda_c\mathcal L_{\mathrm{change}}\\
&+\lambda_n\mathcal L_{\mathrm{nuisance}}
+\lambda_p\mathcal L_{\mathrm{para}}.
\end{split}
\label{eq:total}
\end{equation}
The supervised term counts each verified input once. The four auxiliary weights control distillation, separation, stability, and agreement across paraphrases, respectively.

At inference, the verifier scores $(I,t_i)$ in one pass on the full image (Fig.~\ref{fig:training}(c)). The program, evidence views, and paired edits are needed only during training.

\begingroup
\section{Experiments}
\label{sec:experiments}
\begin{table*}[t]
\centering\small
\renewcommand{\arraystretch}{.94}
\setlength{\tabcolsep}{3pt}
\begin{tabular*}{\textwidth}{@{\extracolsep{\fill}}lrrrrrrrr}
\toprule
 & \multicolumn{4}{c}{Relation group} & \multicolumn{3}{c}{Evaluation split} & \\
\cmidrule(lr){2-5}\cmidrule(lr){6-8}
Model & Spatial & Comparison & Ordering & Graph & \shortstack{New\\scenes} & \shortstack{Unseen\\wording} & \shortstack{Unseen\\compositions} & Overall\\
\midrule
Random Guessing & 50.0 & 50.0 & 50.0 & 50.0 & 50.0 & 50.0 & 50.0 & 50.0\\
\midrule
\resultgroup{9}{Open-source Judges}
Qwen2.5-VL-7B~\cite{qwen25vl} & 57.6 & 59.9 & 57.3 & 59.8 & 60.1 & 56.8 & 59.2 & 58.7\\
InternVL2.5-8B~\cite{internvl25} & 57.7 & 62.2 & 56.4 & 62.3 & 61.6 & 58.4 & 59.2 & 59.7\\
\midrule
\resultgroup{9}{Multimodal Process Reward Models}
MM-PRM-8B~\cite{mmprm} & 50.6 & 50.2 & 50.1 & 50.0 & 50.1 & 50.6 & 50.0 & 50.3\\
URSA-RM-8B~\cite{ursa} & 49.0 & 51.8 & 53.6 & 47.1 & 49.2 & 50.0 & 51.9 & 50.3\\
VisualPRM-8B~\cite{visualprm} & 55.8 & 67.3 & 52.3 & 57.6 & 58.0 & 55.0 & 62.7 & 58.6\\
Qwen-VL-PRM-7B~\cite{ong} & 61.2 & 66.2 & 55.9 & 65.1 & 64.8 & 58.5 & 63.7 & 62.3\\
\textbf{RelPRM-8B} & \textbf{68.4} & \textbf{74.5} & \textbf{63.7} & \textbf{65.6} & \textbf{70.9} & \textbf{64.2} & \textbf{70.1} & \textbf{68.4}\\
\bottomrule
\end{tabular*}
\caption{\textbf{RelProcessBench results} by relation group and evaluation split. Bold denotes the highest score in each column.}
\label{tab:relations}
\end{table*}

\begin{table*}[t]
\centering\small
\renewcommand{\arraystretch}{.94}
\setlength{\tabcolsep}{8pt}
\begin{tabular*}{\textwidth}{@{\extracolsep{\fill}}lrrrrrr}
\toprule
Model & MMMU & MathVision & MathVerse-VO & DynaMath & WeMath & Overall\\
\midrule
Random Guessing & 50.0 & 50.0 & 50.0 & 50.0 & 50.0 & 50.0\\
\midrule
\resultgroup{7}{Proprietary Judges}
GPT-4o~\cite{gpt4o} & 56.3 & 60.2 & 59.7 & 59.0 & 63.3 & 60.3\\
Gemini-2.0-Flash~\cite{geminiflash} & 58.5 & 60.1 & 62.8 & \textbf{66.7} & 58.7 & 62.3\\
GPT-4o-mini & 54.1 & 51.4 & 52.5 & 56.6 & 56.7 & 53.6\\
Doubao-Seed-1.6-Vision & 59.5 & 61.1 & 63.1 & 66.2 & 62.7 & 62.8\\
\midrule
\resultgroup{7}{Open-source Judges}
Qwen2.5-VL-7B~\cite{qwen25vl} & 53.1 & 51.8 & 47.8 & 51.3 & 54.2 & 51.0\\
InternVL2.5-8B~\cite{internvl25} & 47.1 & 45.5 & 47.8 & 50.3 & 50.8 & 48.0\\
Qwen2.5-VL-72B-Instruct~\cite{qwen25vl} & 59.4 & 58.1 & 56.4 & 57.0 & 55.7 & 57.2\\
Qwen3-VL-30B-A3B-Instruct~\cite{qwen3vl} & \textbf{61.3} & 57.5 & 57.4 & 59.0 & 58.8 & 58.2\\
Qwen3-VL-235B-A22B-Instruct~\cite{qwen3vl} & 58.7 & 59.6 & 58.1 & 57.6 & 58.8 & 58.5\\
\midrule
\resultgroup{7}{Multimodal Process Reward Models}
Qwen-VL-PRM-7B~\cite{ong} & 55.8 & 55.7 & 58.8 & 58.3 & 59.8 & 58.6\\
TIM-PRM-8B~\cite{timprm} & 58.3 & 58.3 & 61.9 & 65.9 & 63.9 & 61.7\\
VisualPRM-8B~\cite{visualprm} & 58.5 & 62.1 & 61.0 & 62.7 & 61.8 & 62.0\\
\textbf{RelPRM-8B} & 60.5 & \textbf{66.9} & \textbf{64.7} & \textbf{66.7} & \textbf{66.4} & \textbf{66.2}\\
\bottomrule
\end{tabular*}
\caption{\textbf{VisualProcessBench results.} Bold marks the best score per column.}
\label{tab:vpb}
\end{table*}

\subsection{Experimental setup}
\begingroup
\makeatletter
\renewcommand\paragraph{\@startsection{paragraph}{4}{\z@}{2pt plus 1pt minus 1pt}{-1em}{\normalfont\normalsize\bfseries}}
\makeatother
\paragraph{Training data.}
Starting from VisualPRM-8B, we train on 432,000 RelProcessTrain trajectories and 144,000 VisualPRM400K replay trajectories (3:1). RelProcessTrain supplies program-verified step labels, evidence views, and paired scene edits for the two training mechanisms. Replay retains supervision on general reasoning steps alongside the relational data.

\paragraph{Implementation details.}
We apply LoRA to the language and vision modules, with rank 32, alpha 64, and dropout 0.05, and fully update the projector. AdamW uses $\beta=(0.9,0.999)$ and weight decay $0.05$. Training uses one epoch, an effective batch size of 32 trajectories, BF16, and eight H20 GPUs. We use a learning rate of $10^{-5}$, with cosine decay and 5\% warmup. Loss weights are GT $1.0$, $\lambda_d=\lambda_c=0.1$, and $\lambda_n=\lambda_p=0.05$, with margin $m=1.0$ (Eq.~\eqref{eq:total}).

\paragraph{Benchmarks.}
RPB tests verification across four relation groups and three evaluation splits; VPB~\cite{visualprm} tests broader steps from MMMU~\cite{mmmu}, MathVision~\cite{mathvision}, MathVerse-VO~\cite{mathverse}, DynaMath~\cite{dynamath}, and WeMath~\cite{wemath}. Comparisons include multimodal PRMs and general-purpose multimodal judges. Trajectories are scored step by step in their original order. Each input contains the original image, question, current step, and preceding prefix, excluding later steps, labels, and program answers. RelPRM scores the full image; evidence views and paired edits are absent at evaluation.

\paragraph{Evaluation metrics.}
We evaluate VPB using binary macro-F1 at score $>0.8$ on 24,276 steps from 2,866 trajectories; 2,674 neutral steps remain in context. RPB aggregates family-level balanced accuracy across families and splits, with held-out wording evaluated separately. Appendix~\ref{sec:diagnostic-appendix} details the same-question diagnostic.

\endgroup
\newpage
\subsection{Benchmark results}
\begingroup
\makeatletter
\renewcommand\paragraph{\@startsection{paragraph}{4}{\z@}{2pt plus 1pt minus 1pt}{-1em}{\normalfont\normalsize\bfseries}}
\makeatother
\paragraph{Visual-relation verification.}
RelPRM-8B achieves the highest overall RPB score of 68.4, exceeding VisualPRM-8B by 9.9 points and Qwen-VL-PRM-7B by 6.1 (Table~\ref{tab:relations}). All four relation groups improve over VisualPRM-8B; Spatial and Ordering gain 12.7 and 11.4 points, indicating better verification of positional and ordering claims. Comparison and Graph extend these gains to visual comparisons and graph relations. Together with the matched diagnostic (Fig.~\ref{fig:diagnostic}), these results support better visual-relation error discrimination.

\paragraph{Generalization across evaluation conditions.}
We test transfer to new scenes, expressions, and task compositions. RelPRM improves over VisualPRM-8B by 12.9, 9.3, and 7.4 points on New scenes, Unseen wording, and Unseen compositions, respectively. Gains on independent scenes support transfer across visual instances; gains on unseen wording and compositions extend to expressions and program combinations absent from training. Despite its improvement, unseen wording remains RelPRM's lowest-scoring split, suggesting that aligning new expressions with visual-relation evidence remains challenging.

{\setlength{\abovecaptionskip}{3pt}\renewcommand{\arraystretch}{.94}\begin{table*}[t]
\centering\small
\setlength{\tabcolsep}{4pt}
\begin{tabular*}{\textwidth}{@{\extracolsep{\fill}}lrrrrrrrr}
\toprule
Model & MMMU & MathVista & MathVision & MathVerse-VO & DynaMath & WeMath & LogicVista & Average\\
\midrule
InternVL2.5-8B & 56.2 & 64.5 & 17.0 & 22.8 & 9.4 & 23.5 & 36.0 & 32.8\\
+VisualPRM & 60.2 & 68.5 & \textbf{25.7} & 35.8 & 18.0 & 36.5 & 43.8 & 41.2\\
\textbf{+RelPRM} & \textbf{62.5} & \textbf{69.2} & 25.4 & \textbf{36.4} & \textbf{19.2} & \textbf{41.9} & \textbf{44.3} & \textbf{42.7}\\
\bottomrule
\end{tabular*}
\caption{\textbf{Best-of-8 answer selection across seven benchmarks.} Accuracy (\%); Average is the unweighted dataset mean. Bold denotes the best score in each column.}
\label{tab:bon}
\end{table*}
}
\paragraph{Transfer to broader multimodal step assessment.}
VPB tests whether relation-oriented training benefits broader reasoning steps. RelPRM-8B reaches 66.2, a 4.2-point overall gain over VisualPRM-8B, with improvements on all five source datasets (Table~\ref{tab:vpb}). MathVision and WeMath gain 4.8 and 4.6 points, with improvements also on MMMU, MathVerse-VO, and DynaMath. These gains extend verification benefits beyond RelProcess tasks to broader multimodal reasoning steps. Alongside RPB's verification and generalization results, VPB supports the transfer value of relation-oriented training. Sections~\ref{sec:bon} and~\ref{sec:component-plan} examine answer selection and component contributions.

\endgroup
\newpage
\setlength{\abovecaptionskip}{3pt}
\setlength{\smallskipamount}{1pt}
\makeatletter
\renewcommand\paragraph{\@startsection{paragraph}{4}{\z@}{2pt plus 1pt minus 1pt}{-1em}{\normalfont\normalsize\bfseries}}
\makeatother
\subsection{Best-of-$N$ selection}
\label{sec:bon}
We test whether improved step verification benefits answer selection. InternVL2.5-8B supplies candidates shared by all methods. Evaluation covers the five VPB task sources plus MathVista~\cite{mathvista} and LogicVista~\cite{logicvista} (Table~\ref{tab:bon}).

At $N=8$, RelPRM reaches 42.7\% average accuracy, exceeding VisualPRM by 1.5 points and self-consistency by 2.2. It improves on VisualPRM in six of seven datasets, led by WeMath (5.4 points); MathVision decreases by 0.3 points.
\par\smallskip
\noindent\begin{minipage}{\columnwidth}
\centering
\includegraphics[width=\linewidth]{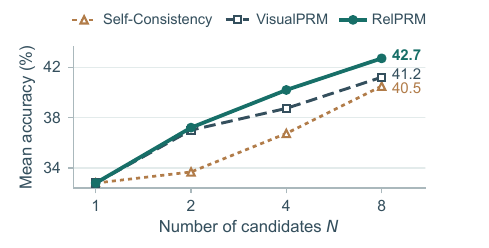}
\captionof{figure}{\textbf{Best-of-$N$ selection.} Seven-dataset mean.}\label{fig:bon}
\end{minipage}\par\smallskip
RelPRM leads VisualPRM at every $N>1$ (Fig.~\ref{fig:bon}), rising from 37.2\% at $N=2$ to 42.7\% at $N=8$. With shared candidates, these gains support the practical value of the verifier for answer selection across pool sizes.

\subsection{Ablation studies}
{\renewcommand{\arraystretch}{.94}\noindent\begin{minipage}{\linewidth}
\centering\small
\setlength{\tabcolsep}{3pt}
\begin{tabular*}{\columnwidth}{@{\extracolsep{\fill}}lrr}
\toprule
Variant & VPB & RPB\\
\midrule
VisualPRM & 62.0 & 58.6\\
\textbf{Full Method (RelPRM)} & \textbf{66.2} & \textbf{68.4}\\
\midrule
GT only & 63.2 & 65.1\\
w/o self-distillation & 64.5 & 67.3\\
w/o consistency training & 65.1 & 66.7\\
\bottomrule
\end{tabular*}
\captionof{table}{\textbf{Framework ablations.} GT only uses verified step labels without either auxiliary objective.}
\label{tab:ablation}
\end{minipage}
}\par\smallskip
\paragraph{Contributions of the two mechanisms.}
Table~\ref{tab:ablation} separates label supervision from the auxiliary training mechanisms. GT-only training gains 1.2 points on VPB and 6.5 on RPB over VisualPRM. Full RelPRM adds another 3.0 and 3.3 points, showing benefits beyond verified-label training.

Removing self-distillation costs 1.7 points on VPB and 1.1 on RPB; removing consistency costs 1.1 and 1.7. In this setting, self-distillation contributes more on VPB and consistency more on RPB.

\par\smallskip{\renewcommand{\arraystretch}{.94}\noindent\begin{minipage}{\linewidth}
\centering\footnotesize
\setlength{\tabcolsep}{3pt}
\begin{tabular*}{\linewidth}{@{\extracolsep{\fill}}p{.73\linewidth}rr@{}}
\toprule
Variant & VPB & RPB\\
\midrule
\textbf{Full RelPRM} & \textbf{66.2} & \textbf{68.4}\\
\resultgroup{3}{Self-distillation ablations}
w/o reliability gate & 65.4 & 67.5\\
Frozen teacher & 65.7 & 67.8\\
Replace distillation with evidence-view supervision & 65.6 & 67.6\\
\resultgroup{3}{Consistency ablations}
w/o truth-change separation & 65.5 & 67.2\\
w/o nuisance stability & 65.8 & 67.7\\
w/o paraphrase-response consistency & 66.1 & 68.2\\
\bottomrule
\end{tabular*}
\captionof{table}{\textbf{Component design ablations.} Each row changes the indicated component of the full framework.}
\label{tab:components}\label{tab:resd-ablation}\label{tab:consistency-ablation}
\end{minipage}
}\par\smallskip
\paragraph{Self-distillation design.}\label{sec:component-plan}
The controls in Table~\ref{tab:components} test teacher reliability, updating, and judgment transfer. Removing the gate costs 0.8 points on VPB and 0.9 on RPB, favoring reliable prediction selection. Freezing the teacher costs 0.5 and 0.6, favoring updated teaching judgments. Replacing distillation with evidence-view supervision costs 0.6 and 0.8. This favors transferring evidence-view predictions to the full-image scorer over adding direct supervision on evidence views.

\paragraph{Consistency design.}
Truth-change separation tests the need to rank true and false versions differently. Its removal costs 0.7 points on VPB and 1.2 on RPB, the largest drop among the consistency terms. Removing nuisance stability costs 0.4 and 0.7, supporting score stability when relevant evidence is preserved. Removing paraphrase-response consistency costs 0.1 and 0.2, a smaller but consistent benefit from aligning edit responses across equivalent wordings. Appendix~\ref{sec:evidence-view-appendix} compares evidence-view construction.

\section{Conclusion}
We present RelPRM, a multimodal process reward modeling method for visual-relation verification. Our automated RelProcess pipeline produces RelProcessTrain and RPB, linking verifiable step labels to visual evidence and controlled relational edits. Relation-evidence self-distillation transfers reliable evidence-assisted judgments to full images. Relation-evidence-driven consistency training constrains scores by step-specific evidence dependencies: responding to truth-changing relational edits, remaining stable under irrelevant edits, and aligning edit responses across paraphrases. Inference requires only the original image and reasoning context. RPB, VPB, and best-of-$N$ results show that training on verifiable visual relations improves relational verification, broader multimodal step assessment, and downstream answer selection.

\endgroup
\clearpage
\section*{LLM Usage Statement}
We used ChatGPT to assist with manuscript drafting, language revision, and experiment execution. The authors take full responsibility for verifying all claims, references, and experimental results.

{\small\bibliographystyle{ieeenat_fullname}\bibliography{references}}
\onecolumn
\begin{center}
{\Large\bfseries Supplementary Material}
\end{center}
\smallskip
\appendix
\section{Same-Question Error-Sensitivity Diagnostic}
\label{sec:diagnostic-appendix}
\paragraph{Construction and scoring.}
We retain the verbatim questions and images of 100 distinct VisualProcessBench (VPB) cases and author a relation step and a later calculation for each task. A relation edit changes one relation phrase without changing numbers; a calculation edit changes only the numeric result, preserving operands and operations. Each target is scored in its own prefix, excluding later steps. Image, question, and pre-target text are fixed within each correct--error pair. VisualPRM-8B and RelPRM receive the same 400 frozen inputs, with all 100 questions retained and no substitutions.

\paragraph{Pairwise ranking.}
For each model and error type, we divide the number of questions satisfying $s_{\mathrm{correct}}>s_{\mathrm{edited}}$ by 100. Ties contribute zero to the numerator and remain in the denominator. Table~\ref{tab:diagnostic-full} gives the paired counts underlying Fig.~\ref{fig:diagnostic}.
\begin{table}[h]
\centering\small
\begin{tabular}{lrr}
\toprule
Model & Calculation wins / pairs & Visual-relation wins / pairs \\
\midrule
VisualPRM-8B & 95/100 & 64/100 \\
RelPRM & 97/100 & 82/100 \\
\bottomrule
\end{tabular}
\caption{\textbf{Strict pairwise ranking on the frozen diagnostic.} Each question contributes one calculation pair and one visual-relation pair.}
\label{tab:diagnostic-full}
\end{table}

\paragraph{Fixed example.}
For the MathVerse triangle (VPB case 2705), the calculation edit replaces only the result of $20\cos(60^\circ)$, changing 10 to 11. The relation edit calls the leg labeled $y$ opposite the marked $60^\circ$ angle. VisualPRM scores the correct and edited calculation 0.818 and 0.037, and the correct and edited relation 0.924 and 0.963. RelPRM gives the corresponding scores 0.807 and 0.059 for calculation and 0.943 and 0.426 for relation.

\clearpage
\section{Evidence-View Construction}
\label{sec:evidence-view-appendix}
\paragraph{Controlled comparison.}
We compare two ways to present the evidence required to judge a reasoning step. \textbf{Object highlighting} emphasizes the relevant objects, while \textbf{distractor masking} hides irrelevant visual content. Both views retain the objects, relations, and dependencies needed for the judgment. The teacher receives the image view, question, reasoning prefix, and current step. Teacher weights and verdict rules are identical across the two conditions, and both are evaluated on the same fixed set of steps with verified labels. Figure~\ref{fig:evidence-example} shows the same step under all three views.

\smallskip
\noindent\begin{minipage}{\linewidth}\centering
\includegraphics[width=.92\linewidth]{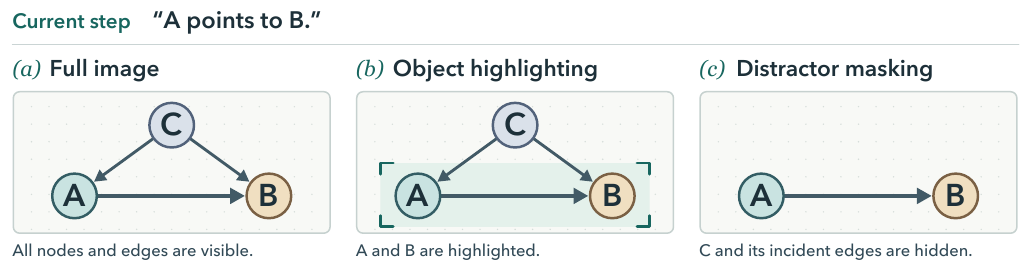}
\captionof{figure}{Evidence views for the step “A points to B.” Both views preserve the directed edge needed to verify the step.}\label{fig:evidence-example}
\end{minipage}\par

\paragraph{Teacher accuracy.}
For each step, we compare the teacher's correct/incorrect prediction $\hat y_j$ with its verified label $y_j$. Over $M$ labeled steps, teacher accuracy is
\begin{equation}
\mathrm{Acc}_{T}=\frac{100}{M}\sum_{j=1}^{M}\ind[\hat y_j=y_j].
\label{eq:teacher-accuracy}
\end{equation}
This measures agreement with step labels under evidence-assisted input; it is distinct from the VPB and RelProcessBench (RPB) scores of the resulting full-image verifier.

\begin{center}
\begin{minipage}{.72\linewidth}
\noindent\begin{minipage}{\linewidth}
\centering\small
\setlength{\tabcolsep}{3pt}
\begin{tabular*}{\linewidth}{@{\extracolsep{\fill}}lrrr@{}}
\toprule
Evidence view & \shortstack{Teacher\\accuracy $\uparrow$} & VPB $\uparrow$ & RPB $\uparrow$\\
\midrule
Object highlighting & 85.0 & 65.9 & 68.0\\
Distractor masking & \textbf{87.0} & \textbf{66.2} & \textbf{68.4}\\
\bottomrule
\end{tabular*}
\captionof{table}{\textbf{Evidence-view comparison.} Teacher accuracy (\%) uses the same fixed set of labeled steps.}
\label{tab:evidence-views}
\end{minipage}

\end{minipage}
\end{center}

\paragraph{Results.}
Distractor masking raises teacher accuracy from 85.0\% to 87.0\%, a gain of 2.0 percentage points (Table~\ref{tab:evidence-views}). The corresponding full-image verifier improves from 65.9 to 66.2 on VPB and from 68.0 to 68.4 on RPB, gains of 0.3 and 0.4 points. Under this comparison, masking therefore provides more accurate teacher judgments and higher downstream verification scores than highlighting. Evidence views are used during training; final benchmark scoring uses the original full image and reasoning context.

\clearpage
\section{RPB Examples by Subtype}
\label{sec:relprocess-examples}
Each panel shows a new-scenes example. We redraw the scene and abbreviate the text for display. Labels and errors come from the benchmark. Paraphrases are illustrative.\par
\noindent\begin{minipage}{\linewidth}\centering
\includegraphics[width=.91\linewidth]{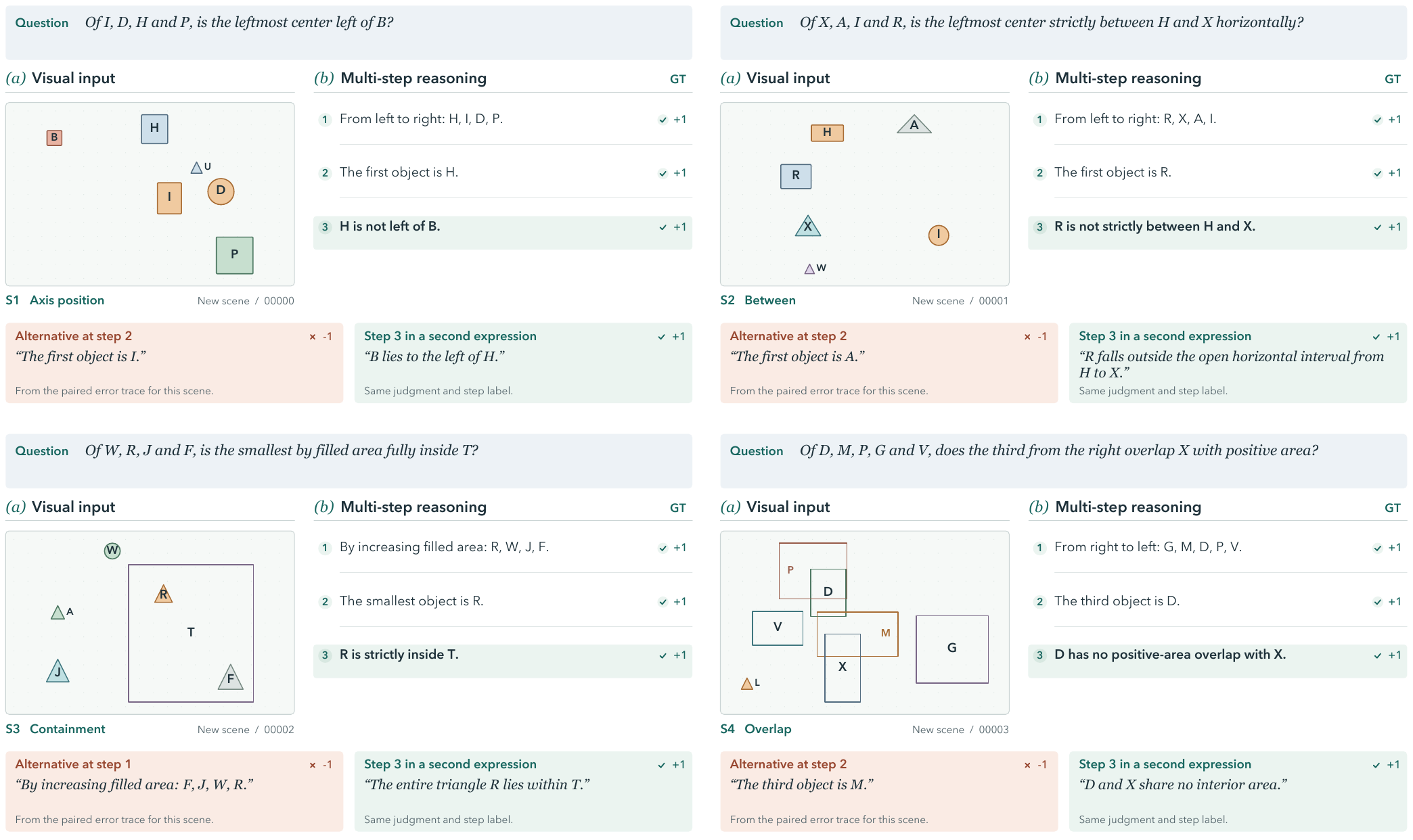}
\captionof{figure}{\textbf{Spatial.} S1 Axis position, S2 Between, S3 Containment, S4 Overlap.}\label{fig:plate-spatial}
\end{minipage}\par
\medskip
\noindent\begin{minipage}{\linewidth}\centering
\includegraphics[width=.91\linewidth]{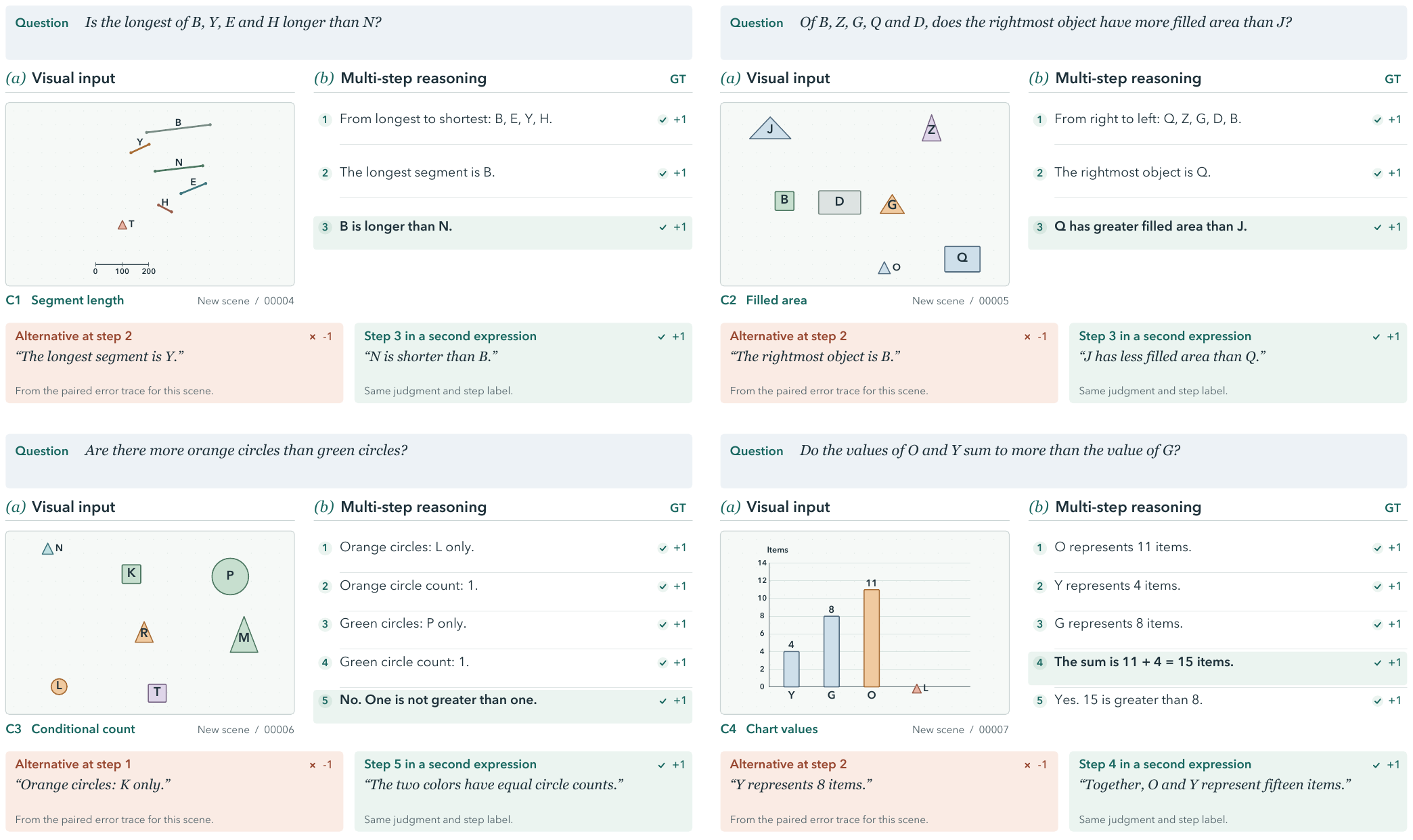}
\captionof{figure}{\textbf{Comparison.} C1 Segment length, C2 Filled area, C3 Conditional count, C4 Chart values.}\label{fig:plate-comparison}
\end{minipage}\par
\clearpage
\noindent\begin{minipage}{\linewidth}\centering
\includegraphics[width=.91\linewidth]{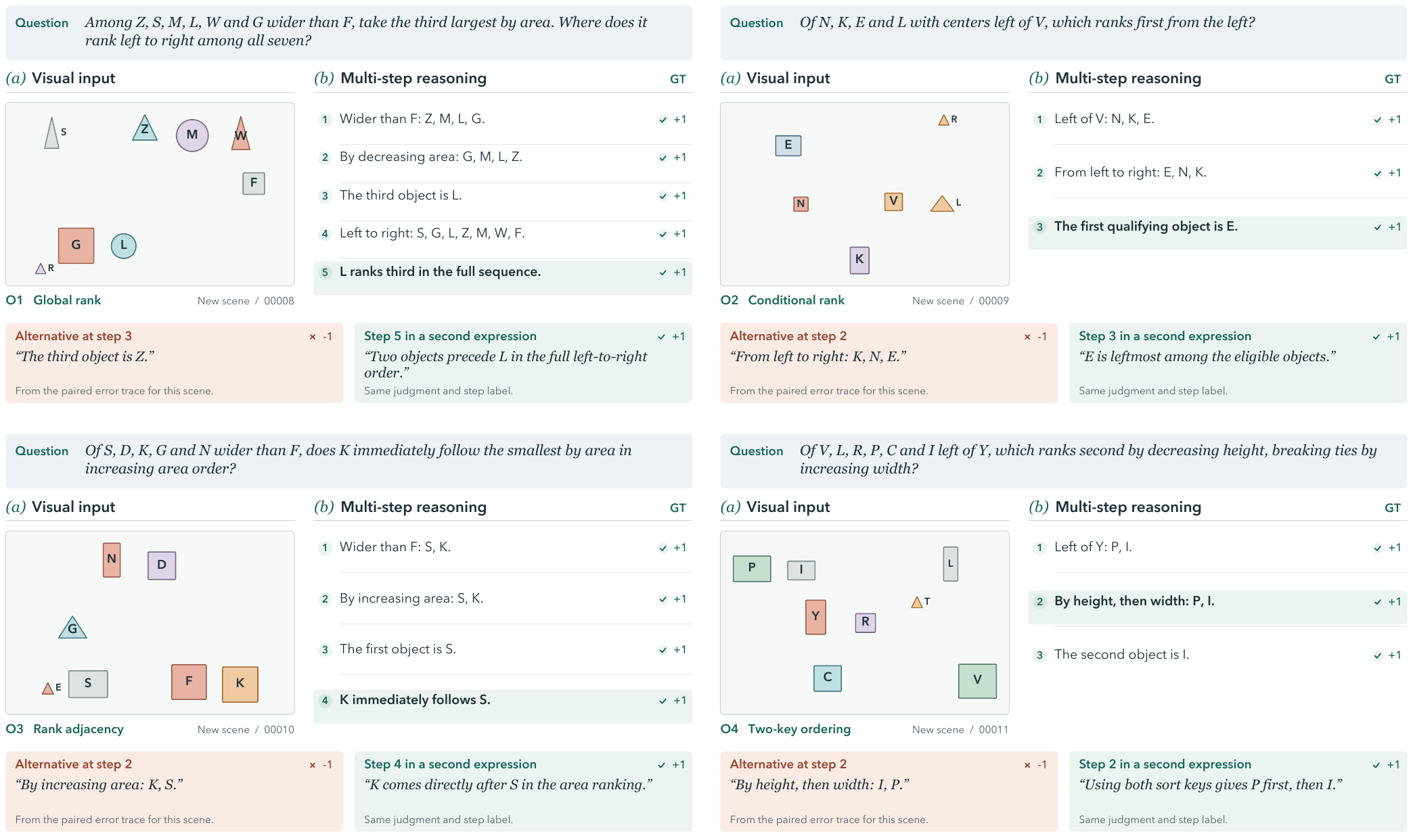}
\captionof{figure}{\textbf{Ordering.} O1 Global rank, O2 Conditional rank, O3 Rank adjacency, O4 Two-key ordering.}\label{fig:plate-ordering}
\end{minipage}\par
\medskip
\noindent\begin{minipage}{\linewidth}\centering
\includegraphics[width=.91\linewidth]{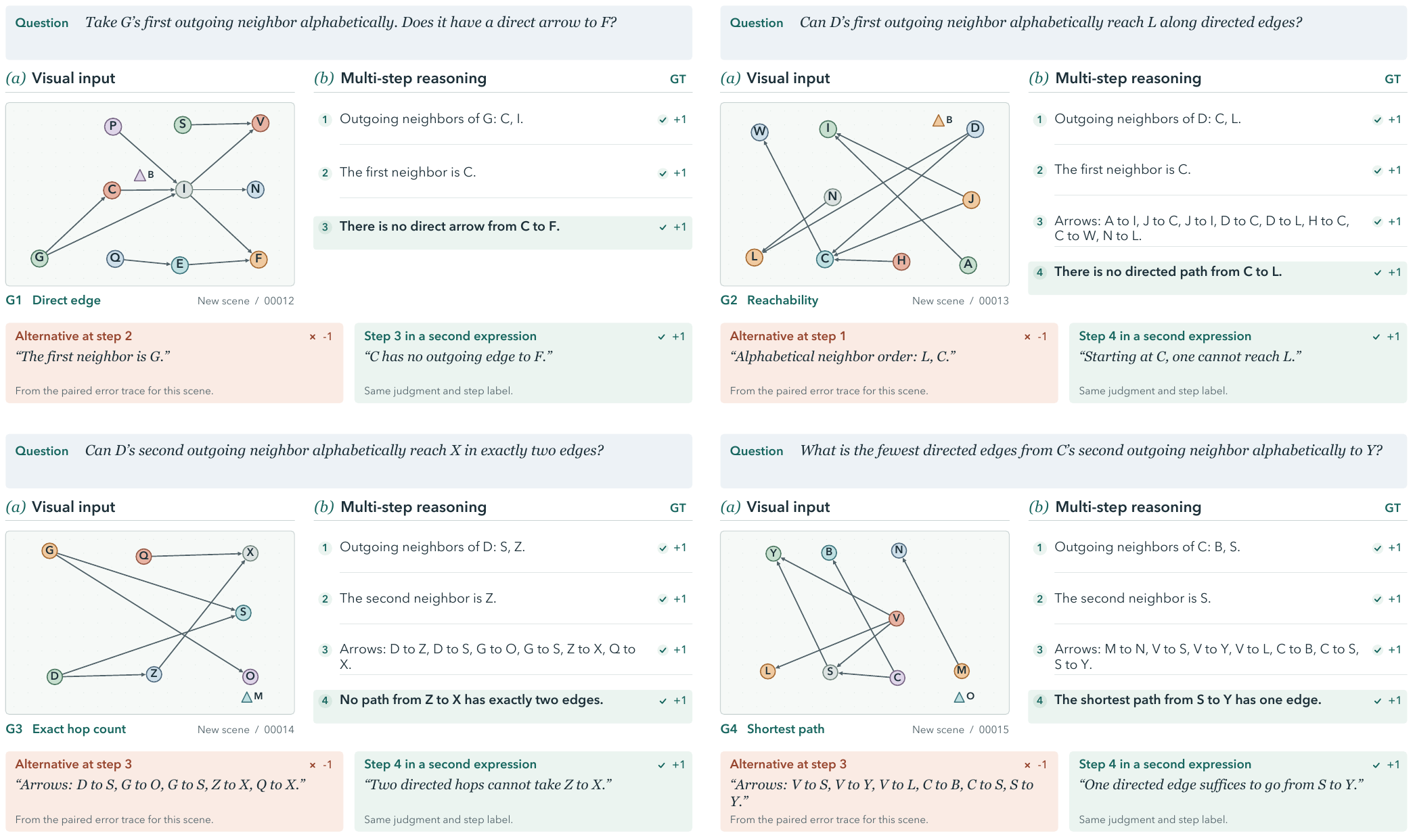}
\captionof{figure}{\textbf{Graph.} G1 Direct edge, G2 Reachability, G3 Exact hop count, G4 Shortest path.}\label{fig:plate-graph}
\end{minipage}\par

\clearpage
\section{Relational Training Across Difficulty Levels}
\label{sec:difficulty-analysis}
We compare VisualPRM-8B, its relation-data-trained variant (+ RelProcessTrain), and the full RelPRM model on the Simple, Medium, and Hard subsets of RPB (Table~\ref{tab:difficulty-analysis}). This comparison examines whether gains from relation-data training persist as evaluation difficulty increases.

\begin{center}
\begin{minipage}{.82\linewidth}
\centering\small
\setlength{\tabcolsep}{8pt}
\begin{tabular*}{\linewidth}{@{\extracolsep{\fill}}lrrrr}
\toprule
Model & Simple & Medium & Hard & Overall\\
\midrule
VisualPRM-8B & 59.9 & 58.9 & 56.8 & 58.6\\
+ RelProcessTrain & 68.2 & 66.3 & 60.8 & 65.1\\
\textbf{RelPRM-8B} & \textbf{70.3} & \textbf{69.2} & \textbf{65.8} & \textbf{68.4}\\
\bottomrule
\end{tabular*}
\captionof{table}{\textbf{Effect of relational training across RPB difficulty levels.} The second row augments VisualPRM-8B with relation-data training. Bold denotes the best score in each column.}
\label{tab:difficulty-analysis}
\end{minipage}
\end{center}

Relation-data training improves all three subsets over VisualPRM-8B, but the gain decreases from 8.3 points on Simple to 4.0 on Hard. Its score falls from 68.2 to 60.8 between these subsets, a 7.4-point gap. Full RelPRM improves over this variant by 2.1, 2.9, and 5.0 points on Simple, Medium, and Hard, respectively, with the largest additional gain on Hard. The Simple--Hard gap narrows to 4.5 points. These results support the motivation for supplementing relational labels with evidence-assisted judgment transfer and evidence-driven consistency: the full framework retains more of the benefit on harder examples, although a difficulty gap remains.

\end{document}